%% file: main.tex
\pdfoutput=1

\documentclass[11pt]{article}

\usepackage[final]{acl}

\usepackage{hyperref}
\usepackage{times}
\usepackage{latexsym}
\usepackage{url}
\usepackage{listings}
\usepackage{xcolor}
\usepackage{multirow}
\usepackage{booktabs}
\usepackage{pgfplots}
\pgfplotsset{compat=newest}
\usepackage{pgf-umlsd}
\usepackage{tikz}
\usepackage{lipsum}
\usetikzlibrary{shapes.geometric, arrows, positioning, fit}
\usepackage{amsmath} 
\usepackage{pgfplots}
\pgfplotsset{compat=1.18}
\usepackage{caption}
\usepackage{pdflscape}
\usetikzlibrary{positioning, arrows.meta}
\usepackage{rotating}

\usepackage[T1]{fontenc}

\usepackage[utf8]{inputenc}

\usepackage{microtype}

\usepackage{inconsolata}

\usepackage{graphicx}
\usepackage{natbib}

\title{SandboxAQ's submission to MRL 2024 Shared Task on Multi-lingual Multi-task Information Retrieval}

\author{
  \begin{tabular}{c@{\hspace{2em}}c}
    Isidora Chara Tourni & Sayontan Ghosh \\
    {\normalfont\texttt{isidora.tourni@gmail.com}} & {\normalfont\texttt{sayontan.ghosh@gmail.com}} \\
 \\[2ex]
    Brenda Miao & Constantijn van der Poel \\
    {\normalfont\texttt{brenda.miao@gmail.com}} & {\normalfont\texttt{constantijn@sandboxquantum.com}}
  \end{tabular}
}

\begin{document}
\maketitle
\input{abstract}

\input{intro}

\input{related}

\input{dataset}
\input{qa}

\input{ner}
\input{methods}
\input{results}

\input{conclusions}

\newpage
\bibliography{custom}


\label{sec:bibtex}



\appendix

\input{appendix}

\end{document}

%% file: abstract.tex
\begin{abstract}
This paper explores the problems of Question Answering (QA) and Named Entity Recognition (NER) in five diverse languages. We tested five Large Language Models with various prompting methods, including zero-shot, chain-of-thought reasoning, and translation techniques. Our results show that while some models consistently outperform others, their effectiveness varies significantly across tasks and languages. We saw that advanced prompting techniques generally improved QA performance but had mixed results for NER; and we observed that language difficulty patterns differed between tasks. Our findings highlight the need for task-specific approaches in multilingual NLP and suggest that current models may develop different linguistic competencies for different tasks.
\end{abstract}

%% file: intro.tex
\section{Introduction}
The need for effective multilingual information retrieval systems has never been more pressing. Despite rapid advancements in the fields of Natural Language Processing (NLP) and Neural Machine Translation (NMT), creating systems that can operate across diverse languages and tasks remains a significant challenge \citep{hu2022language, 10.1145/3626772.3657943, yang2024distillation, guo2024steering, elmahdy2024synergistic}. We addressed this need as part of our participation in the MRL 2024 Shared Task on Multi-lingual Multi-task Information Retrieval\footnote{\url{https://sigtyp.github.io/st2024-mrl.html}}, focusing on two critical tasks: Question Answering (QA) and Named Entity Recognition (NER). QA \citep{pandya2021question} involves answering questions based on given contexts in various languages; while in NER \citep{keraghel2024survey,pakhale2023comprehensive}, the goal is to identify and classify named entities (e.g., person names, organizations, locations) in text across multiple languages.

With recent advances in the zero-shot and instruction-following capabilities of Large Language Models (LLMs), we examined how well existing state-of-the-art models can handle these tasks. More importantly, we also explore whether advanced prompting strategies, which have been shown to significantly boost performance across various natural language and general reasoning benchmarks \citep{toma2024wanglab,d2024context,shi2024generate,battle2024unreasonable}, could also improve the multilingual performance of these LLMs in low-resource settings. We firstly saw how models perform in a zero-shot setting and if they can handle QA and NER in multiple languages with no task-specific training and continued with prompting techniques applied through DSPy\footnote{\url{https://github.com/stanfordnlp/dspy}} \cite{khattab2022demonstrate, khattab2023dspy}. We also explored whether pivoting through English improves performance for the examined low-resource languages, and, finally, combined these approaches, using DSPy-enhanced prompts with English translation of the input.

The datasets used in this task presented several intriguing challenges that pushed the boundaries of current model capabilities. These include handling dialectal variations, while working with low-resource languages that have limited NLP resources; exploring multi-task learning through the combination of QA and NER tasks; and using complex cultural contexts embedded in the questions and passages. Despite these challenges, these datasets offer a unique opportunity to evaluate the robustness and flexibility of LLMs across diverse languages and tasks, and contribute to the advancement of truly multilingual NLP systems, capable of understanding and processing language in its rich and culturally nuanced forms.

In summary, this work answers the following questions:
\begin{itemize}
\itemsep0em 
\item How well do state-of-the-art language models perform in multilingual, multi-task setups?
\item Can prompting techniques significantly boost performance across languages?
\item Is there an advantage to English translation as an intermediate step to multilingual tasks?
\end{itemize}

Our experiments show that while we were able to achieve very high results in both tasks using zero-shot approaches, there was variability in performance across models, methods, and languages. The complex interactions between model architectures, prompting strategies, and linguistic features suggest that future advancements may require more nuanced, language-specific approaches rather than one-size-fits-all solutions. The effectiveness of translation and advanced prompting techniques varies greatly, requiring careful consideration when applying these methods to different languages and models.
Our code can be found at \url{https://github.com/sandbox-quantum/mrl2024}.

%% file: related.tex
\section{Related Work}
\label{sec:related}

\subsection*{Multilingual LLMs}

Researchers have explored multilingual capabilities of state-of-the-art LLMs through various approaches, using multilingual benchmarks \citep{zhang2023m3exam}, and have worked towards improving multilingual performance through techniques such as cross-lingual prompting \citep{huang2023not}, phonemic transcription integration \citep{nguyen2023enhancing}, and soft prompting with multilingual verbalizers \citep{li2023enhancing}.

In recent works, \citet{marchisio2024understanding} investigated LLMs' language confusion, introducing the Language Confusion Benchmark (LCB) to evaluate inconsistencies across 15 diverse languages, while \citet{lu2024llamax} developed LLaMAX, to enhance translation capabilities for over 100 languages through multilingual continual pre-training. \citet{aryabumi2024aya} expanded upon the Aya model \citet{ustun2024aya} to present Aya 23, which supports 23 languages and expands its capabilities to about half the world's population.
Also, \citet{zhao2024large} proposed a three-stage multilingual workflow hypothesis and introduced Parallel Language-specific Neuron Detection (PLND) to identify language-specific neurons. Finally, \citet{hengle2024multilingual} created the MultiLingual Needle in a Haystack test, revealing LLMs' sensitivity to language and position of information in long contexts.

While studies enhanced our understanding of how LLMs process multilingual inputs, they also highlight areas for improvement in multilingual capabilities.

\subsection*{NMT with LLMs}

The landscape of NMT has been significantly transformed by LLMs. Earlier studies by \citet{robinson-etal-2023-chatgpt} and \citet{zhu2023multilingual} highlighted LLMs' competitiveness with state-of-the-art NMT systems for high-resource languages, while noting limitations in low-resource scenarios. Recently, \citet{enis2024llm} showed that Claude 3 Opus demonstrates strong capabilities across resource levels, often surpassing specialized systems.  \citet{yan2024gpt} found that GPT-4's performance is comparable to junior translators but still lags behind more experienced ones, with quality gradually becoming worse from resource-rich to resource-poor language pairs. \citet{he2024exploring} explored mimicking human translation strategies with LLMs, showing that incorporating preparatory steps like keyword and topic analysis can improve translation quality and reduce errors. Adding to this, \citet{hasan2024large} highlighted significant performance disparities between English and low-resource South Asian languages, finding that while GPT-4 consistently outperformed other LLMs, all models showed superior performance in English compared to languages like Bangla, Hindi, and Urdu. 

These insights suggest that while LLMs are making significant strides in translation tasks, their performance varies considerably across language pairs and resource levels, showing the potential and current limitations of LLM-based translation.

\subsection*{Multilingual QA}

Multilingual and Cross-lingual QA have emerged as critical areas of NLP research, driven by the advance of LLMs and the creation of diverse multilingual QA datasets. Benchmarks such as TYDI QA \citep{clark2020tydi}, MLQA \citep{lewis2019mlqa}, XQuAD \citep{artetxe2019cross}, and MKQA \citep{longpre2021mkqa} have been instrumental in enabling evaluation across a wide range of languages, representing different language families and typologies.
Most of these multilingual datasets, however, are still limited to evaluation on benchmarking datasets and not yet tested in real-world settings.

Various works have focused on generating synthetic multilingual QA data for training downstream QA models \citep{riabi2020synthetic,shakeri2020towards}. Notable contributions include PAXQA \citep{li2023paxqa}, generating cross-lingual QA datasets by leveraging parallel corpora and without requiring annotated non-English QA data. More recently, \citet{englander2024m2qa} introduced M2QA, a multi-domain multilingual QA benchmark of Indian languages.

For QA, zero-shot transfer learning has been studied extensively; models are typically fine-tuned on English data and then applied directly to other languages, leveraging cross-lingual capabilities of models like mBERT \citep{devlin2018bert} and XLM-R \citep{conneau2019unsupervised}. 
CORA \citep{asai2020xor} introduces a unified model for cross-lingual retrieval and multilingual QA, using NMT for data augmentation, either by translating English QA pairs into target languages or by translating non-English queries to English at inference time. \citet{duan2021bridging} assemble triples from different languages and use a knowledge injection strategy, via link prediction, to enrich a model's multilinguality.

Latest significant contributions include QAMELEON \citep{agrawal2023qameleon}, which uses prompt-tuning to generate synthetic QA data with as few as 5 examples per language.
Also, \citet{carrino2023promoting} focused on the challenging task of the question being in a different language from the context in QA data; they improve cross-lingual QA transfer by utilizing cross-lingual sampling and advanced self-distillation, to enhance a high-performing multilingual model trained on a large-scale dataset, and complemented by a few thousand aligned QA examples across languages.

\subsection*{Multilingual NER}
Recent advancements in multilingual NER have focused on overcoming challenges such as limited annotated data for low-resource languages and performance gaps across languages. Approaches like Retrieval Augmented Generation \citep{tan2023damo}, contrastive learning \citep{mo2024mcl}, and knowledge integration \citep{fetahu2023semeval} have shown promising results. The U-RaNER system \citep{tan2023damo} leverages external knowledge sources and multiple retrieval strategies, while MCL-NER \citep{mo2024mcl} employs multi-view contrastive learning to align cross-lingual representations. Data augmentation techniques, including translation \citep{wu2020unitrans} and code-switching \citep{mo2024mcl}, have been used to create synthetic training data for low-resource languages. Self-training on unlabeled target language data has also proven effective \citep{wu2020single,mo2024mcl}.

Transformer-based models with knowledge integration consistently perform well, 
with domain adaptation appearing more important than language adaptation \citep{fetahu2023semeval,tan2023damo,Balluff2024,kulkarni-etal-2023-towards}. A MultiCoNER\footnote{\url{https://multiconer.github.io/}} system by \citet{fetahu2023semeval} highlighted the effectiveness of ensemble strategies and external knowledge for handling complex entity types. \citet{kulkarni-etal-2023-towards} showed that when including domain-specific adaptations, a unified model can perform close to multiple monolingual models with fewer parameters. \citet{mayhew2023universal} introduced UNER, gold-standard NER benchmarks across languages. 

Despite these advancements, fine-grained entity types and noisy input data remain significant challenges. With top systems achieving F1 scores above 85\%, in benchmarks like XTREME \citep{siddhant2020xtreme} and UNER \citep{mayhew2023universal}, there is still room for improvement in low-resource languages, noisy text scenarios, and fine-grained entity recognition.

%% file: dataset.tex
\section{Datasets}
\label{sec:datasets}
\input{data_sizes}

Our study utilized datasets provided by the MRL 2024 Shared Task, found in the XTREME-UP repository\footnote{\url{https://github.com/google-research/xtreme-up}}, covering a wide range of topics, like history, science, technology, and local culture. Example snippets are given in Appendix \ref{sec:appendix_examples}.

\subsection*{Languages}

The datasets cover five languages, representing a diverse range of Language families\footnote{\url{https://en.wikipedia.org/wiki/Language_family}} and resource availability. This selection includes both high- and low-resource languages, allowing for a comprehensive evaluation of multilingual capabilities:

\begin{itemize}
\itemsep0em 
    \item Igbo (\textbf{IG}) (Niger-Congo family) 
    \item Swiss German / Alemannic German (\textbf{ALS}) (Germanic family)
    \item Turkish (\textbf{TR}) (Turkic family)
    \item Azerbaijani (\textbf{AZ}) (Turkic family)
    \item Yoruba (\textbf{YO}) (Niger-Congo family)
\end{itemize}

\subsection*{QA Dataset}

The QA dataset consists of question-answer pairs along with corresponding text passages in each language.
The data is in CSV format with columns for \textit{annotation\_id}, \textit{text}, \textit{question}, \textit{answer options (A, B, C, D)}, and \textit{label}. The task is multiple-choice question answering, where systems must select the right answer from four options.

\subsection*{NER Dataset}

The NER dataset consists of text samples in each of the above languages, except IG, annotated with named entities.
The data is in CoNLL\footnote{\url{https://www.clips.uantwerpen.be/conll2003/ner/}} format, with each token on a separate line. It uses the BIO (Beginning, Inside, Outside) tagging scheme, and three types of entities are annotated:
    \begin{itemize}
    \itemsep0em 
        \item PER: Person names
        \item ORG: Organization names
        \item LOC: Location names
    \end{itemize}

\subsection*{Dataset Splits}
Data sizes for each task are given in Table \ref{tab:language-datasets}. For both tasks, the data is divided into two splits:

\begin{itemize}
\itemsep0em 
    \item Validation set: Used for hyperparameter tuning and preliminary evaluation.
    \item Test set: Unlabeled data, used for final evaluation and reporting of results.
\end{itemize}

%% file: data_sizes.tex
\begin{table*}[t]
\centering
\begin{tabular}{lcccc}
\hline
\textbf{Language} & \textbf{QA Val Set Size} & \textbf{QA Test Set Size} & \textbf{NER Val Set Size} & \textbf{NER Test Set Size}\\
\hline
Igbo (\textbf{IG}) & 201 & 748& -&11813\\
Swiss German (\textbf{ALS}) & 201 & 651 &8893&11778\\
Turkish (\textbf{TR}) & 196 & 148&7490&11889\\
Azerbaijani (\textbf{AZ}) & 201 & 291&7902&8326 \\
Yoruba (\textbf{YO}) & 201 & 673 &4268&11795\\
\hline
\end{tabular}
\caption{Validation and Test Set Sizes in number of examples.}
\label{tab:language-datasets}
\end{table*}

%% file: qa.tex
\begin{table*}[t]
\centering
\small
\begin{tabular}{llccccc|cc}
\hline
Model & QA Method & ALS & YO & AZ & IG & TR & Avg \\
\hline
\multirow{6}{*}{gpt-4o} & Simple QA & 94.10 & 94.10 & 98.00 & 98.20 & 96.92 & \textbf{96.26 $\pm$ 2.03} \\
 & English Translation & 92.50 & 90.00 & 97.50 & 97.00 & 95.38 & \textbf{94.48 $\pm$ 3.17} \\
 & DSPy CoT & 91.50 & 91.50 & 92.00 & 91.50 & 91.79 & 91.66 $\pm$ 0.23 \\
 & DSPy CoT + Translation & 89.50 & 90.00 & 89.52 & 90.00 & 89.74 & 89.75 $\pm$ 0.25 \\
 & DSPy-ReAct & 91.60 & 88.90 & 94.80 & 96.70 & 96.82 & 93.76 $\pm$ 3.44 \\
 & DSPy-ReAct + Translation & 97.04 & 95.50 & 97.00 & 95.02 & 96.92 & \textbf{96.30 $\pm$ 0.96} \\
\hline
\multirow{6}{*}{gpt-4-turbo} & Simple QA & 88.20 & 90.10 & 96.63 & 95.50 & 95.18 & 93.12 $\pm$ 3.73 \\
 & English Translation & 97.00 & 93.00 & 99.00 & 96.50 & 96.23 & \textbf{96.35 $\pm$ 2.16} \\
 & DSPy CoT & 96.50 & 97.00 & 98.00 & \textbf{98.50} & 97.59 & \textbf{97.52 $\pm$ 0.79} \\
 & DSPy CoT + Translation & 97.50 & \textbf{99.00} & 98.50 & 96.50 & 96.92 & \textbf{97.68 $\pm$ 1.05} \\
 & DSPy-ReAct & 95.50 & 95.00 & 97.50 & 96.50 & 97.44 & 96.39 $\pm$ 1.13  \\
 & DSPy-ReAct + Translation & 97.50 & 97.84 & 98.00 & 98.01 & 97.86 & \textbf{97.84 $\pm$ 0.21}\\
\hline
\multirow{6}{*}{claude-3.5-sonnet} & Simple QA & 92.50 & 94.50 & \textbf{99.50} & 93.00 & 96.93 & 95.29 $\pm$ 2.92\\
 & English Translation & 95.00 & 96.00 & 94.00 & 94.50 & 94.87 & \textbf{94.87 $\pm$ 0.74} \\
 & DSPy CoT & 93.00 & 91.50 & 92.00 & 91.00 & 92.82 & 92.06 $\pm$ 0.85  \\
 & DSPy CoT + Translation & 92.00 & 92.00 & 94.50 & 95.00 & 96.41 & \textbf{93.98 $\pm$ 1.94}\\
 & DSPy-ReAct & 92.00 & 90.00 & 98.00 & 94.00 & 97.44 & \textbf{94.29 $\pm$ 3.44} \\
 & DSPy-ReAct + Translation & 93.00 & 98.50 & 99.00 & 94.50 & 96.93 & \textbf{96.39 $\pm$ 2.58} \\
\hline
\multirow{6}{*}{Aya-23-8B} & Simple QA & 42.00 & 59.00 & 43.00 & 28.50 & 57.95 & 46.09 $\pm$ 12.68 \\
 & English Translation & 42.50 & 64.00 & 48.00 & 45.50 & 66.67 & 53.33 $\pm$ 11.17 \\
 & DSPy CoT & 95.50 & 92.50 & 92.50 & 92.52 & 93.85 & 93.37 $\pm$ 1.32 \\
 & DSPy CoT + Translation & 94.00 & 94.10 & 94.02 & 94.20 & 93.85 & 94.03 $\pm$ 0.13 \\
 & DSPy-ReAct & 94.03 & 94.10 & 94.00 & 94.00 & 93.97 & 94.02 $\pm$  0.05 \\
 & DSPy-ReAct + Translation & 95.50 & 98.60 & 98.51 & \textbf{98.52} & 97.89 & \textbf{97.80 $\pm$ 1.32} \\
\hline
\multirow{6}{*}{Aya-23-35B} & Simple QA & 59.50 & 58.50 & 74.50 & 65.50 & 68.11 & 65.22 $\pm$ 6.57 \\
 & English Translation & 62.50 & 56.50 & 73.00 & 64.00 & 81.54 & 67.51  $\pm$ 9.82 \\
 & DSPy CoT & \textbf{98.50} & 98.40 & 98.51 & 97.00 & 92.31 & \textbf{96.94  $\pm$ 2.67} \\
 & DSPy CoT + Translation & 94.00 & 94.02 & 94.00 & 94.10 & 93.85 & 93.99  $\pm$ 0.09\\
 & DSPy-ReAct & 94.00 & 97.00 & 98.51 & \textbf{98.50} & \textbf{98.46} & \textbf{97.29 $\pm$ 1.95} \\
 & DSPy-ReAct + Translation & 98.00 & 98.50 & 98.50 & 97.00 & 92.31 & \textbf{96.86 $\pm$  2.62}\\
\hline
Average & & 88.53  & 89.66 & 91.35 & 89.38 & 91.83 & 90.15 $\pm$ 13.70 \\

\hline
\end{tabular}
\caption{Model Validation Set Performance Comparison on QA. The best results (\% Accuracy) per language are in bold. The rightmost column shows the average across all languages, with the highest performing method for each model in bold. The final row shows average performance across models and methods for each language. }
\label{tab:qa}
\end{table*}

%% file: ner.tex
\begin{table*}[t]
\centering
\tiny
\begin{tabular}{llcccccccccccc|c}
\hline
 & & \multicolumn{3}{c}{ALS} & \multicolumn{3}{c}{YO} & \multicolumn{3}{c}{AZ} & \multicolumn{3}{c}{TR} & \multicolumn{1}{c}{Avg} \\
Model & NER Method & P & R & F1 & P & R & F1 & P & R & F1 & P & R & F1 & F1 \\
\hline
\multirow{6}{*}{gpt-4o} 
 & Simple NER & \textbf{94.10} & \textbf{94.10} & \textbf{92.01} & 90.66 & 91.55 & 90.57 & 89.61 & 90.41 & 89.25 & \textbf{92.27} & \textbf{92.56} & \textbf{92.23} & \textbf{91.02}\\
 & English Translation & 92.15 & 92.60 & 91.90 & 91.31 & 91.77 & 91.04 & 89.40 & 89.67 & 88.50 & 91.43 & 90.95 & 90.56 & 90.50 \\
 & DSPy CoT & 88.73 & 90.34 & 88.32 & \textbf{91.94} & \textbf{92.44} & \textbf{91.32} & 91.18 & \textbf{92.13} & 90.70 & 87.90 & 89.15 & 86.88 & 89.31 \\
 & DSPy CoT + Translation & 87.04 & 88.11 & 83.59 & 86.74 & 88.62 & 85.69 & 87.28 & 89.89 & 86.05 & 85.21 & 87.32 & 82.63 & 84.49 \\
 & DSPy ReAct & 88.73 & 90.34 & 88.32 & \textbf{91.94} & \textbf{92.44} & 91.32 & \textbf{91.19} & 92.10 & \textbf{90.71} & 87.99 & 89.15 & 86.80 & 89.29 \\
 & DSPy ReAct + Translation & 87.04 & 88.11 & 83.59 & 86.74 & 88.62 & 85.69 & 87.28 & 89.89 & 86.05 & 85.21 & 87.32 & 82.63 & 84.49 \\
\hline
\multirow{6}{*}{gpt-4-turbo} 
 & Simple NER & 88.95 & 90.43 & 87.88 & 86.59 & 88.62 & 85.91 & 86.46 & 87.89 & 84.62 & 86.46 & 87.89 & 84.62 & 85.76 \\
 & English Translation & \textbf{90.06} & \textbf{90.85} & \textbf{89.78} & \textbf{90.07} & 90.86 & 89.78 & 88.15 & 89.07 & 87.30 & \textbf{88.85} & \textbf{89.30} & \textbf{87.96} & \textbf{88.71} \\
 & DSPy CoT & 85.91 & 88.58 & 85.26 & 90.02 & \textbf{91.62} & 90.32 & \textbf{89.32} & \textbf{90.80} & 88.10 & 85.70 & 87.93 & 84.19 & 86.97 \\
 & DSPy CoT + Translation & 88.57 & 90.04 & 85.97 & 85.60 & 87.81 & 82.83 & 84.24 & 86.75 & 81.29 & 86.87 & 88.14 & 84.80 & 83.72 \\
 & DSPy ReAct & 85.91 & 88.58 & 85.26 & 90.02 & \textbf{91.62} & \textbf{90.33} & 89.30 & 90.78 & \textbf{88.11} & 85.70 & 87.93 & 84.19 & 86.97 \\
 & DSPy ReAct + Translation & 85.60 & 87.81 & 82.83 & 86.87 & 88.14 & 84.80 & 88.57 & 90.04 & 85.97 & 84.24 & 86.75 & 81.29 & 83.72 \\
 \hline
\multirow{6}{*}{claude-3.5-sonnet} 
 & Simple NER & \textbf{94.02} & \textbf{94.18} & \textbf{93.85} & 93.05 & 92.29 & 92.61 & 94.02 & 94.18 & 93.85 & \textbf{89.60} & \textbf{90.10} & \textbf{89.07} & \textbf{92.35} \\
 & English Translation & 86.93 & 88.68 & 85.25 & 87.12 & 90.28 & 87.89 & 90.92 & 91.51 & 89.28 & 88.09 & 88.65 & 85.60 & 87.01 \\
 & DSPy CoT & 92.30 & 92.68 & 92.68 & \textbf{93.59} & \textbf{92.96} & \textbf{93.21} & 94.40 & 94.48 & 94.18 & 89.23 & 89.85 & 88.71 & 92.20 \\
 & DSPy CoT + Translation & 86.51 & 88.60 & 85.04 & 88.78 & 89.65 & 87.98 & 87.06 & 89.96 & 86.43 & 86.29 & 87.56 & 83.38 & 85.71 \\
 & DSPy ReAct & 87.98 & 89.41 & 87.62 & 91.05 & 90.66 & 90.70 & \textbf{96.27} & \textbf{95.31} & \textbf{95.31} & 89.02 & 89.54& 88.29 & 90.48 \\
 & DSPy ReAct + Translation & 87.41 & 89.26 & 86.66 & 89.10 & 89.60 & 89.03 & 90.27 & 91.21 & 89.11 & 87.14 & 88.02 & 84.48 & 87.32 \\
\hline
\multirow{6}{*}{Aya-23-8B} 
 & Simple NER & 80.36 & \textbf{89.65} & \textbf{84.75} & 76.33 & \textbf{87.37} & \textbf{81.47} & 74.82 & 86.50 & 80.23 & 72.13 & 84.93 & 78.01 & 81.12 \\
 & English Translation & 80.36 & \textbf{89.65} & \textbf{84.75} & 74.82 & 86.50 & 80.23 & 72.13 & 84.93 & 78.01 & 76.33 & \textbf{87.37} & \textbf{81.47} & 81.12 \\
 & DSPy CoT & 85.14 & 87.60 & 84.43 & \textbf{79.30} & 83.72 & 79.42 & 86.19 & 89.13 & 86.35 & 80.12 & 86.17 & 80.90 & 82.78 \\
 & DSPy CoT + Translation & 84.76 & 87.73 & 84.49 & 78.82 & 83.60 & 79.44 & 86.45 & \textbf{89.18} & \textbf{86.58} & 80.72 & 86.23 & 81.15 & \textbf{82.92} \\
 & DSPy ReAct & \textbf{85.19} & 87.61 & 84.28 & \textbf{79.30} & 83.63 & 79.34 & \textbf{86.51} & 88.87 & 86.26 & 80.24 & 86.25 & 81.07 & 82.74 \\
 & DSPy ReAct + Translation & 85.17 & 87.85 & 84.57 & 79.26 & 83.68 & 79.70 & 85.22 & 88.96 & 85.73 & \textbf{80.86} & 86.25 & 81.20 & 82.80 \\
\hline
\multirow{6}{*}{Aya-23-35B} 
 & Simple NER & 76.33 & 87.37 & 81.47 & 72.13 & \textbf{84.93} & 78.01 & 80.36 & \textbf{89.65} & 84.75 & 74.82 & \textbf{86.50} & 80.23 & 81.12 \\
 & English Translation & 76.33 & 87.37 & 81.47 & 72.13 & \textbf{84.93} & 78.01 & 80.36 & \textbf{89.65} & 84.75 & 74.82 & 86.49 & 80.23 & 81.12 \\
 & DSPy CoT & 76.32 & 87.36 & 81.47 & 72.12 & 84.92 & 78.00 & 80.36 & 89.64 & 84.75 & 74.81 & 86.45 & 80.23 & 81.11 \\
 & DSPy CoT + Translation & 84.98 & 87.79 & 84.92 & 82.58 & 84.68 & 82.69 & 85.15 & 88.87 & \textbf{85.86} & 79.84 & 85.66 & 80.88 & 83.59 \\
 & DSPy ReAct & \textbf{86.29} & \textbf{88.14} & \textbf{84.93} & \textbf{83.45} & 84.80 & \textbf{82.92} & \textbf{86.74} & 89.49 & 85.73 & \textbf{82.04} & 86.21 & \textbf{81.15} & \textbf{83.68} \\
 & DSPy ReAct + Translation & 83.69 & 87.55 & 83.70 & 82.70 & 84.73 & 82.77 & 84.08 & 88.52 & 85.32 & 78.86 & 85.98 & 80.56 & 83.09 \\
\hline
Average & & 86.17 & 89.25 & 86.20 & 84.88 & 88.07 & 85.37 & 86.78 & 89.75 & 86.94 & 83.71 & 87.86 & 84.24 & 85.69 \\
\hline
\end{tabular}
\caption{Named Entity Recognition (NER) Validation Results: Precision (P), Recall (R), and F1 Scores. Best results per language and model are in bold. The rightmost column shows the average F1 score across all languages.}
\label{tab:ner}
\end{table*}

%% file: methods.tex
\section{Methods}
\label{sec:methods}

We proposed a systematic approach to evaluate the performance of state-of-the-art LLMs on multilingual QA and NER. We used a variety of prompting strategies to assess and enhance the models' capabilities. This section details our experimental setup, the models used, the evaluation methods and metrics, and the prompting techniques applied.

Due to the recent surge in advanced LLMs and their associated costs, combined with our own time and budget constraints, we chose to focus on a select group of models.
We evaluated the following five LLMs, representing different architectures and training approaches:

\begin{itemize}
\itemsep0em 
\item gpt-4o \citep{achiam2023gpt}
\item gpt-4-turbo \citep{achiam2023gpt}
\item claude-3.5-sonnet\footnote{\url{https://docs.anthropic.com/en/docs/welcome}}
\item Aya-23-8B\footnote{\url{https://huggingface.co/CohereForAI/aya-23-8B}} \cite{aryabumi2024aya}
\item Aya-23-35B\footnote{\url{https://huggingface.co/CohereForAI/aya-23-35B}} \cite{aryabumi2024aya}
\end{itemize}

We worked towards developing robust solutions for both tasks, using the provided datasets.
Our work proposed six different evaluation strategies to assess the models' performance:

\begin{itemize}
\itemsep0em 
\item Zero-shot Learning: We test the models' ability to perform NER and QA tasks without any task-specific fine-tuning or examples.

\item English Translation: We prompt the model to first translate the input text to English before processing.

\item DSPy Chain of Thought (DSPy CoT): We use the DSPy library to implement chain-of-thought reasoning, encouraging models to break down their reasoning process into steps.

\item DSPy Chain-of-Thought + Translation: Data is translated to English before applying the Chain of Thought approach.

\item DSPy ReAct: We apply the Reasoning and Acting (ReAct) approach from DSPy, which combines step-by-step reasoning with simulated actions.

\item DSPy ReAct + Translation: Data is translated to English before applying DSPy ReAct.

\end{itemize}

DSPy \citep{khattab2022demonstrate,khattab2023dspy} represents a paradigm shift in prompt engineering, moving towards a more algorithmic and systematic approach.
Its power lies in its optimizers, that automatically refine prompts or adjust model weights to improve performance on a given metric. It optimizes each component for different models. However, this library and the optimizations provided have not been extensively tested on multilingual tasks.

DSPy ChainOfThought enhances a task's input-output structure by incorporating a reasoning step. It adds a `rationale' component between the input and final output, allowing the LLM to articulate its thought process. This helps break down complex problems into more manageable steps, potentially improving problem-solving and making the model's reasoning more transparent. DSPy ReAct, on the other hand, implements the Reasoning and Acting approach. This module alternates between thinking steps and concrete actions, such as retrieving information or using external tools. ReAct offers a structured method for combining analytical steps with practical actions or API calls, and enables more sophisticated problem-solving that can integrate external data or resources.

For each model and evaluation strategy, we followed these steps (a flowchart of the process can be found in Appendix \ref{sec:appendix_process}):

\begin{enumerate}
\itemsep0em 
\item Data Preparation: We preprocessed the dataset for each file and language, ensuring consistent formatting across all inputs.

\item Prompt Engineering: We designed prompts for each task and strategy.

\item Model Inference: We ran each model on the prepared data using the appropriate prompts, incorporating DSPy techniques where applicable.

\item Output Processing: We post-processed the model outputs.

\item Evaluation: We computed the following performance metrics for each task:

\begin{itemize}
\itemsep0em 
    \item For QA: Accuracy
    \item For NER: Precision, Recall, F1 score
\end{itemize}

\item Cross-lingual Analysis: We compared performance across different languages to assess each model's multilingual capabilities.

\item Cross-model and methods Comparison: We analyzed how different model architectures and methods perform relative to each other across tasks and languages.
\end{enumerate}

We used the official APIs for OpenAI's GPT-4\footnote{\url{https://openai.com/index/openai-api/}} and Anthropic's claude models\footnote{\url{https://www.anthropic.com/api}}, and Ollama versions for the aya Language Models\footnote{\url{https://ollama.com/library/aya}}, namely \textit{aya:8b} as the 8B version, and \textit{aya:35b-23-q2\_K} as the quantized 35B version.
For DSPy, we used the DSPy library\footnote{\url{https://dspy-docs.vercel.app/}}, and specifically dspy-Chain-of-Thought\footnote{\url{https://dspy-docs.vercel.app/api/modules/ChainOfThought}} and dspy-ReAct\footnote{\url{https://dspy-docs.vercel.app/api/modules/ReAct}}.
Our prompts for all experiments are provided in Appendix \ref{sec:appendix_prompts}.

All experiments were ran on an AWS g5.8xlarge instance, with 32 vCPUs, 128 GiB in memory and an NVIDIA A10G GPU.
We initially explored temperature values in the range of [0,1] for OpenAI and Anthropic models, using a fixed seed. Our preliminary tests revealed negligible variation in model outputs across this temperature range. Consequently, we standardized the temperature value to 0 for all subsequent experiments to ensure consistency and reproducibility of results.

GPT-4o in a simple zero-shot setup was used to create predictions for the competition submissions, for both QA and NER.

%% file: results.tex
\section{Results \& Discussion}
\label{sec:results}
Results for QA and NER are given in Tables \ref{tab:qa} and \ref{tab:ner}, respectively. Figures showing accuracy for our best performing models in both tasks are provided in 
Appendix \ref{sec:appendix_figures} 
 Tables \ref{tab:qa_delta} and \ref{tab:ner_delta} in the Appendix \ref{sec:appendix_tables} provide information on model performance compared to baseline Simple QA or NER averaged across languages. 

\subsection*{QA}

Our analysis of multilingual QA revealed important insights. Baseline differences were seen in model performance across languages, with AZ, TR and IG tending to show higher scores than YO or ALS. On average, GPT-4-turbo and GPT-4o tended to outperform other models, particularly in zero-shot settings. DSPy-enhanced prompts boosted performance for all models, and especially for the open source Aya models. Both the 8B and 35B parameter Aya models showed comparable average performance to the other models with the use of DSPy-ReAct with English Translation, with accuracies of 97.80\% and 96.86\%, respectively. In fact, \textbf{DSPy ReAct with English translation showed the highest average improvements over Simple QA across four out of the five languages}, with improvements ranging from an average of 16.10\% for TR to 27.83\% for ALS. 

Despite comparable average performances across models with DSPy prompting, the effect of prompting varied significantly when stratified by individual languages and model. For GPT-4-turbo, DSPy CoT improved scores by an average of $4.72\%$ across languages, with notable improvements in YO (from $90.10\%$ to $97.00\%$) and ALS (from $88.20\%$ to $96.50\%$). In contrast, DSPy CoT decreased Claude-3.5-sonnet's performance by an average of $4.62\%$, with the most significant drop in AZ ($99.50\%$ to $92.00\%$). Interestingly, GPT-4o saw a decrease in performance using DsPy CoT, as well as DSPy ReAct and DsPy CoT with English Translation compared to simple QA, but saw a $1.11\%$ improvement when using ReAct with English translation.

We further examined the impact of translating text to English before processing. Translation alone tended to maintain or improve performance compared to simple QA, with the largest average improvements of 4.72\% occurring for the TR language. Again, when looking at individual models and languages, English translation had varying effects. Translation improved performance for GPT-4-turbo in simple QA tasks by an average of $3.46\%$, with the most significant improvement in ALS ($88.20\%$ to $97.00\%$). Claude-3.5-sonnet showed mixed results with translation alone, experiencing a slight average decrease of $0.91\%$ compared to simple QA. Combining DSPy CoT with Translation only improved performance compared to CoT alone. However, adding Translation to DSPy ReAct boosted performance across all languages, except TR. YO saw the largest average gains in accuracy of 5.15\% with DSPy ReAct + Translation compared to DSPy ReAct alone.

Our results additionally revealed a nuanced relationship between model size and performance; raw parameter count isn't the sole determinant of multilingual and multi-task capabilities. The Aya models lagged significantly behind in simple QA, yet showed a clear performance increase from 8B to 35B parameters, with average simple QA improvements of $15.32\%$ across languages. \textbf{On average, prompting methods and pivoting through English translations tended to significantly improve the performance of these smaller open source models.} The most dramatic improvement was seen in AZ, where performance increased from $28.50\%$ with Simple QA to $98.50\%$ with DSPy-ReAct + Translation for the Aya-23-8B model. Additionally, while Claude-3.5-sonnet and GPT-4-turbo often showed strong performance, the Aya models, which are assumed to be much smaller, were able to match or outperform these models with some DSPy prompting approaches. 

Models that excelled in simple QA performed relatively well across all prompting strategies. However, the improvements from advanced techniques were not uniform, indicating that the choice of QA method should be tailored to the specific language and model combination for optimal results.

 \subsection*{NER}

Our results on multilingual NER across five models and four languages reveal distinct patterns in LLMs' multilingual processing capabilities (\ref{tab:ner}). Models tended to perform better on ALS (86.20\% average microF1) and AZ (86.94\%)  compared to YO (85.37\%) and TR (84.24\%) on NER, whereas models tended to show higher accuracy for TR and AZ in QA, as previously described. 

GPT-4o and Claude-3.5-sonnet consistently outperformed other models in simple NER. GPT-4o achieved an average F1 score of $91.02\%$ across all languages, with its best performance in ALS ($92.01\%$) and TR ($92.23\%$). Claude-3.5-sonnet averaged $92.35\%$, showing particularly strong results in ALS and AZ (both $93.85\%$). This performance surpassed that of GPT-4-turbo ($85.76\%$ average F1) and both Aya models ($81.12\%$ average F1).

\textbf{Unlike QA, model size did not directly correlate with performance for NER}. The difference in performance between Aya-23-35B and Aya-23-8B was minimal in simple NER tasks, with the smaller model showing comparable or better performance for some languages. For example, in ALS, Aya-23-8B had an average F1 score of $84.75\%$  versus $81.47\%$ for Aya-23-35B. 

A\textbf{dvanced prompting techniques were also less effective in improving average performance for NER. }GPT-4o saw a slight average decrease of $1.59\%$ in F1 score with DSPy CoT across all languages, with the most significant drop in ALS ($92.01\%$ to $88.32\%$). In contrast, Claude-3.5-sonnet benefited from DSPy CoT, with an average increase of $0.38\%$ in F1 score.

For DSPy ReAct, GPT-4-turbo showed mixed results, with an average improvement of $0.95\%$ and particularly effective in YO ($85.91\%$ to $90.33\%$). The Aya models, saw more consistent improvements with DSPy techniques, with Aya-23-35B achieving an average F1 score increase of $2.62\%$ using DSPy ReAct. For Claude-3.5-sonnet, DSPy ReAct improved performance on AZ by $1.46\%$ but led to a $6.23\%$ decrease in ALS. Similarly, GPT-4-turbo saw a $4.42\%$ improvement in YO with DSPy ReAct but a $2.62\%$ decline in ALS. 

Translating input to English also tended to hurt performance, even when combined with DSPy prompting approaches. GPT-4-turbo saw an average improvement of $2.34\%$, with significant gains in YO ($85.91\%$ to $89.78\%$) but .PT-4o showed an average decrease of $0.96\%$ in F1 score with translation across all languages. Claude-3.5-sonnet also experienced performance drops, with an average decrease of $4.45\%$ in F1 score, most notable in ALS ($93.85\%$ to $85.25\%$). 


\subsection*{Discussion}
Our analysis of QA and NER tasks across multiple languages reveals differences in how LLMs perform on these tasks. GPT-4o,  GPT-4-turbo, and Claude-3.5-sonnet showed the best performance on QA but GPT-4-turbo lagged behind in NER, though still outperforming the Aya models.

The impact of advanced techniques differed between tasks. DSPy-enhanced prompts generally improved QA performance, particularly for the Aya models. However, in NER, the impact was more varied, with fewer models and languages showing improvements. Translation effects also diverged; in QA, they varied widely with GPT-4-turbo often improving, while in NER, translation generally led to decreased performance, especially for GPT-4o and Claude-3.5-sonnet.

Language difficulty patterns also differed between tasks. In QA, AZ and TR had higher scores, while in NER, ALS and AZ outperformed YO and TR. This suggests that language difficulty can vary between tasks, with performance differences generally more pronounced in NER. Model selection also affected capabilities on different languages, with Claude-3.5-sonnet showed consistent performance across languages, while GPT-4 models had more variability.

Our findings highlight the importance of task-specific optimization in multilingual NLP. Techniques that boost performance in one task may be less effective in another, and models may show different linguistic competencies across tasks.

%% file: conclusions.tex
\section{Conclusions}
\label{sec:conclusions}

Our work on multilingual QA and NER and the variability in results across languages and tasks indicates that no single approach served as a universal solution across all scenarios.

The effectiveness of techniques like DSPy and translation varied between models and tasks. We challenged the assumption that larger models always perform better, suggesting that factors such as model architecture and prompt development play a significant role. Our results show that models specifically designed for multilingual tasks did not consistently outperform general-purpose LLMs, suggesting that specialized architectures may not be necessary for effective multilingual processing.
A key advantage of our approach is achieving high scores (often above $95\%$ for QA and $90\%$ for NER) with minimal system building. This shows the potential for efficient, yet highly effective multilingual NLP systems through simple utilization of pretrained models.

Our research also suggests the potential of using open-source LLMs beyond the Aya models, including LLama 3\footnote{\url{https://llama.meta.com/}}, Mixtral \citep{jiang2023mistral7b, jiang2024mixtralexperts}, and Gemma 2 \citep{team2024gemma}. These models, which have often demonstrated performance comparable to closed-source alternatives, may facilitate new research and practical applications for working with limited resources in future studies.

Other future research may focus on approaches that better handle the diverse requirements of different languages and tasks. These efforts, combined with the use of language specific datasets, or texts in languages of related language families to the languages we examine, for model finetuning, will be crucial in building truly effective and resource-efficient systems.

%% file: appendix.tex
\clearpage
\section{Appendix}
\label{sec:appendix}

\subsection{Dataset examples}
\label{sec:appendix_examples}
\textbf{NER}:
\begin{verbatim}
Dr -X- _ O
Ludwig -X- _ B-PER
Wilhelm -X- _ I-PER
Erhard -X- _ I-PER
( -X- _ O
* -X- _ O
4. -X- _ O
Februar -X- _ O
1897 -X- _ O
z -X- _ O
Fürth -X- _ B-LOC
; -X- _ O
† -X- _ O
5. -X- _ O
Mai -X- _ O
1977 -X- _ O
z -X- _ O
Bonn -X- _ B-LOC
) -X- _ O
isch -X- _ O
e -X- _ O
dytsche -X- _ O
Bolitker -X- _ O
un -X- _ O
Wirtschaftswisseschaftler -X- _ O
gsii. -X- _ O
\end{verbatim}

\hspace{-4mm}\textbf{QA}:
\begin{verbatim}
annotation_id,text,question,A,B,C,D,label
34920690,Di Messenischi Chrieg bizachnet 
meriri Chrieg wo i de Antiki stattgfunde hend 
und zo de Underwerffig vo Messenie dör 
d Spartaner gfüert hend. Datierig und 
d Aazaal vo de Chrieg isch i de Antiki und i de 
moderne Forschig nöd aihaitlich.,Ih wellere 
Ziitperiode hend di Messenische Chrieg 
stattgfunde?,
Ih de Antiki. ,Ih Messenie.,
Ih Sparta.,Im Mittelalter.,A
36318343,"De Pausanias datiert di Messenische 
Chrieg ufs Joor gnau ahand vo de Olympiade. Doch 
werd hütt agnoo, as sini Daate öppe vierzg Joor 
z früüe sind.",Wie gross isch de unterschied 
zwüschet de date vom Pausanias und dem was hüt 
agnoh wird?,De Pausanius het sich bide Date 
um eppe 40 Jahr gege unde vertueh.,Gar kein 
Unterschied,Genau richtig.,Zwänzg Johr. ,A

\end{verbatim}

\subsection{Prompts}
\label{sec:appendix_prompts}
\begin{itemize}
\itemsep0em 
\item  \underline{QA prompt:} 
\newline
"For each of the examples given, read the text and question. Choose the best answer from the options A, B, C, or D provided below. Respond with ONLY the letter (A, B, C, or D) of the correct answer. If the text does not contain the exact information to answer the question, pick the answer closer to the truth. If none of the answers are a good pick, if there are answers or options missing or if you cannot make a decision, pick a letter from A, B, C, D at random."

\item \underline{English Translation prompt:} 
\newline "Translate the following text to English:"

\item \underline{QA with English translation prompt:} 
\newline
"Translate the following text to English: \newline For each of the examples given, read the text and question. Choose the best answer from the options A, B, C, or D provided below. Respond with ONLY the letter (A, B, C, or D) of the correct answer. If the text does not contain the exact information to answer the question, pick the answer closer to the truth. If none of the answers are a good pick, if there are answers or options missing or if you cannot make a decision, pick a letter from A, B, C, D at random."

\item \underline{NER prompt:} 
\newline
"Extract named entities from the text using ONLY the following tags, where \newline B- indicates Beginning /  \newline 
I- indicates Inside the word\newline
B-/I-PER: Person, 
B-/I-ORG: Organization, 
B-/I-LOC: Location, 
O: Non-entity\newline
Provide each entity and its tag in the format 'Entity (TAG)'.
Give a tag for every word in the text."

\item \underline{NER with English translation prompt:} 
\newline
"Translate the following text to English: Extract named entities from the text using ONLY the following tags, where \newline B- indicates Beginning / \newline I- indicates Inside the word\newline 
B-/I-PER: Person, 
B-/I-ORG: Organization, 
B-/I-LOC: Location, 
O: Non-entity\newline
Provide each entity and its tag in the format 'Entity (TAG)'.
Give a tag for every word in the text."

\end{itemize}



\newpage
\subsection{Method flowchart}
\label{sec:appendix_process}
\input{process}  
\newpage

\newpage

\input{plot_qa_ner}

\newpage
\begin{table*}[!t]
\vspace{2mm}
\subsection{Model Performance by Language}
\label{sec:appendix_tables}

\centering
\small

\begin{tabular}{l| c c c c c}

 & \textbf{ALS} & \textbf{YO} & \textbf{AZ} & \textbf{IG} & \textbf{TR} \\
\hline
\textbf{Simple QA} & -- & --  & --  & --  & --  \\
\textbf{English Translation} & 3.51\% & 0.83\%& -0.03\%& 4.41\%& 4.72\%\\
\textbf{DSPy CoT} & 26.23\%& 18.85\%& 14.91\%& 23.59\%& 12.83\%\\
\textbf{DSPy CoT + Translation} & 24.10\%& 18.40\%& 14.31\%& 23.40\%& 13.41\%\\
\textbf{DSPy ReAct} & 24.14\%& 17.36\%& 17.29\%& 26.00\%& 16.63\%\\
\textbf{DSPy ReAct + English Translation} & 27.83\%& 23.41\%& 19.28\%& 26.88\%& 16.10\%\\

\end{tabular}
\caption{Change in Average Validation Set Performance over Simple QA by Language \& Prompting Method.}
\label{tab:qa_delta}
\end{table*}

\begin{table*}[!h]
\centering
\small

\begin{tabular}{l |c c c c}

 & \textbf{ALS} & \textbf{YO} & \textbf{AZ} & \textbf{TR} \\
\hline
\textbf{Simple NER} & -- & --  & --  & --  \\
\textbf{English Translation} & -1.55\%& -0.38\%& -1.12\%& 0.39\%\\
\textbf{DSPy CoT} & -1.77\%& 0.86\%& 2.63\%& -0.77\%\\
\textbf{DSPy CoT + Translation} & -3.63\%& -2.32\%& -1.50\%& -2.67\%\\
\textbf{DSPy ReAct} & -2.17\%& 1.41\%& 3.10\%& -0.63\%\\
\textbf{DSPy ReAct + English Translation} & -4.23\%& -1.54\%& -0.12\%& -3.30\%\\

\end{tabular}
\caption{Change in Average Validation Set Performance over Simple NER by Language \& Prompting Method.}
\label{tab:ner_delta}
\end{table*}

%% file: process.tex
\usetikzlibrary{shapes.geometric, arrows, positioning, fit}


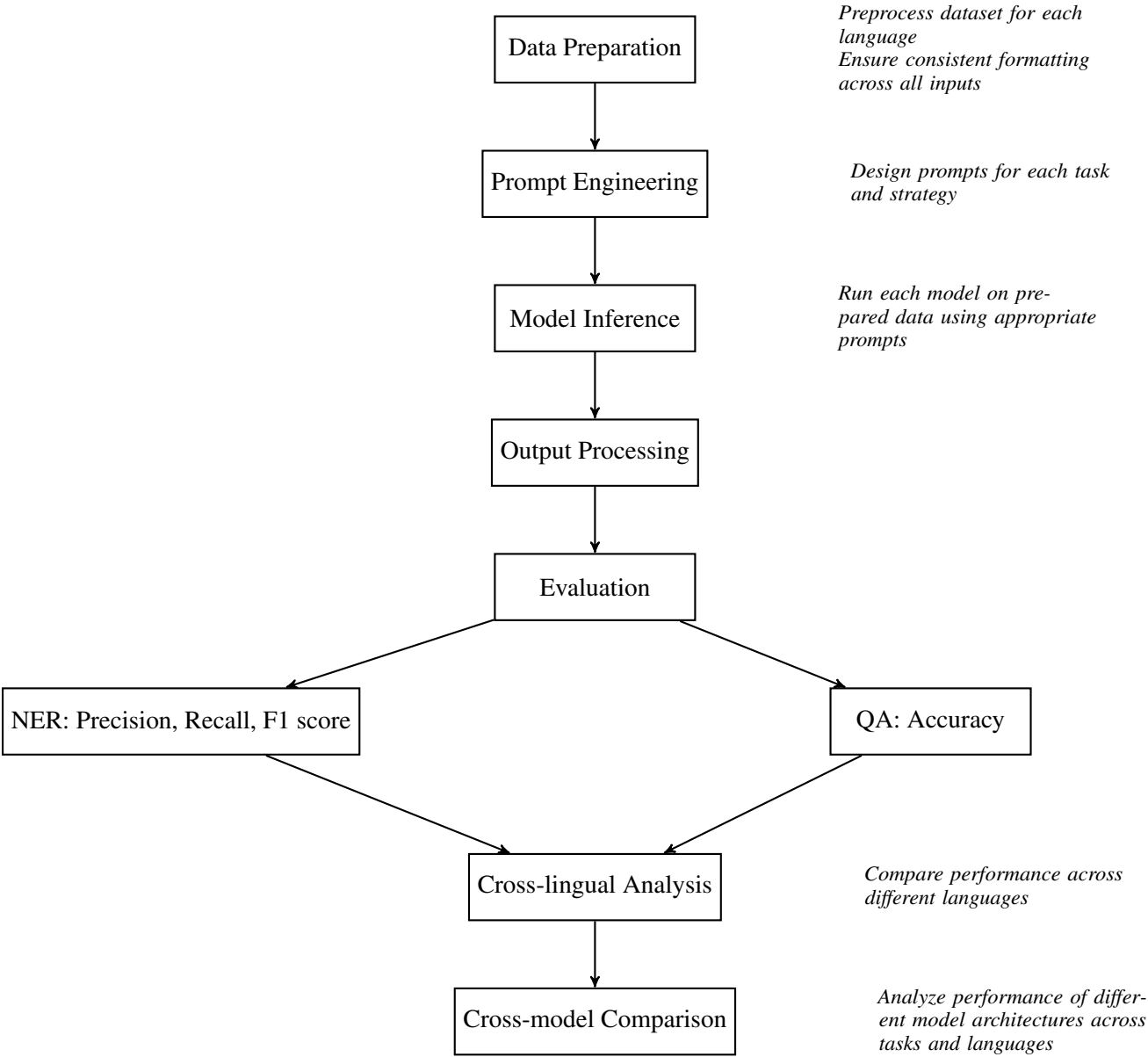
\begin{figure}[!h]
\centering
\begin{tikzpicture}[
    auto,
    node distance = 1cm and 2cm,
    thick,
    process/.style={
      rectangle,
      minimum width=3cm,
      minimum height=1cm,
      text centered,
      draw=black
    },
    arrow/.style={->, >=stealth'},
    note/.style={text width=4cm, font=\small\itshape}
]

\node[process] (start) {Data Preparation};
\node[process, below=of start] (prompt) {Prompt Engineering};
\node[process, below=of prompt] (infer) {Model Inference};
\node[process, below=of infer] (output) {Output Processing};
\node[process, below=of output] (eval) {Evaluation};
\node[process, below left=1cm and 2cm of eval] (ner) {NER: Precision, Recall, F1 score};
\node[process, below right=1cm and 2cm of eval] (qa) {QA: Accuracy};
\node[process, below=of eval, yshift=-25mm] (cross) {Cross-lingual Analysis};
\node[process, below=of cross] (compare) {Cross-model Comparison};

\draw[arrow] (start) -- (prompt);
\draw[arrow] (prompt) -- (infer);
\draw[arrow] (infer) -- (output);
\draw[arrow] (output) -- (eval);
\draw[arrow] (eval) -- (ner);
\draw[arrow] (eval) -- (qa);
\draw[arrow] (ner) -- (cross);
\draw[arrow] (qa) -- (cross);
\draw[arrow] (cross) -- (compare);

\node[note, right=of start] {Preprocess dataset for each language\\Ensure consistent formatting across all inputs};
\node[note, right=of prompt] {Design prompts for each task and strategy};
\node[note, right=of infer] {Run each model on prepared data using appropriate prompts};
\node[note, right=of cross] {Compare performance across different languages};
\node[note, right=of compare] {Analyze performance of different model architectures across tasks and languages};


\end{tikzpicture}
\caption{Method Flowchart}
\label{fig:evaluation-process}
\end{figure}

%% file: plot_qa_ner.tex
\begin{figure*}[!t]
\vspace{2mm}
\subsection{QA and NER performance}
\label{sec:appendix_figures}

\captionsetup{width=.8\textwidth}
\label{fig:comprehensive-qa-performance-line}
\centering
\begin{tikzpicture}
\begin{axis}[
    width=15cm,
    height=10cm,
    xlabel={Language},
    ylabel={Accuracy (\%)},
    title={QA Performance: gpt-4o vs gpt-4-turbo vs claude-3.5-sonnet},
    xticklabels={ALS,YO,AZ,IG,TR},
    xtick={1,2,3,4,5},
    ymin=75,
    ymax=100,
    ymajorgrids=true,
    grid style=dashed,
    legend style={
        at={(0.5,1.25)},
        legend columns=3,
        anchor=south,
        font=\footnotesize
    },
]
\addplot[color=purple,mark=*] coordinates {
    (1,94.10) (2,94.10) (3,98.00) (4,98.20) (5,96.92)
};
\addplot[color=purple,mark=square] coordinates {
    (1,92.50) (2,90.00) (3,97.50) (4,97.00) (5,95.38)
};
\addplot[color=purple,mark=triangle] coordinates {
    (1,91.50) (2,91.50) (3,92.00) (4,91.50) (5,91.79)
};
\addplot[color=purple,mark=diamond] coordinates {
    (1,89.50) (2,90.00) (3,89.52) (4,90.00) (5,89.74)
};
\addplot[color=purple,mark=pentagon] coordinates {
    (1,91.60) (2,88.90) (3,94.80) (4,96.70) (5,96.82)
};
\addplot[color=purple,mark=star] coordinates {
    (1,97.04) (2,95.50) (3,97.00) (4,95.02) (5,96.92)
};
\addplot[color=orange,mark=*] coordinates {
    (1,88.20) (2,90.10) (3,96.63) (4,95.50) (5,95.18)
};
\addplot[color=orange,mark=square] coordinates {
    (1,97.00) (2,93.00) (3,99.00) (4,96.50) (5,96.23)
};
\addplot[color=orange,mark=triangle] coordinates {
    (1,96.50) (2,97.00) (3,98.00) (4,98.50) (5,97.59)
};
\addplot[color=orange,mark=diamond] coordinates {
    (1,97.50) (2,99.00) (3,98.50) (4,96.50) (5,96.92)
};
\addplot[color=orange,mark=pentagon] coordinates {
    (1,95.50) (2,95.00) (3,97.50) (4,96.50) (5,97.44)
};
\addplot[color=orange,mark=star] coordinates {
    (1,97.50) (2,97.84) (3,98.00) (4,98.01) (5,97.86)
};
\addplot[color=teal,mark=*] coordinates {
    (1,92.50) (2,94.50) (3,99.50) (4,93.00) (5,96.93)
}; 
\addplot[color=teal,mark=square] coordinates {
    (1,95.00) (2,96.00) (3,94.00) (4,94.50) (5,94.87)
}; 
\addplot[color=teal,mark=triangle] coordinates {
    (1,93.00) (2,91.50) (3,92.00) (4,91.00) (5,92.82)
}; 
\addplot[color=teal,mark=diamond] coordinates {
    (1,92.00) (2,92.00) (3,94.50) (4,95.00) (5,96.41)
}; 
\addplot[color=teal,mark=pentagon] coordinates {
    (1,92.00) (2,90.00) (3,98.00) (4,94.00) (5,97.44)
}; 
\addplot[color=teal,mark=star] coordinates {
    (1,93.00) (2,98.50) (3,99.00) (4,94.50) (5,96.93)
};
\legend{
  gpt-4o,
  gpt-4o + T,
  gpt-4o DSPy CoT,
  gpt-4o DSPy CoT + T,
  gpt-4o DSPy ReAct,
  gpt-4o DSPy ReAct + T,
  gpt-4-turbo,
  gpt-4-turbo + T,
  gpt-4-turbo DSPy CoT,
  gpt-4-turbo DSPy CoT + T,
  gpt-4-turbo DSPy ReAct,
  gpt-4-turbo DSPy ReAct + T,
  claude-3.5-sonnet,
  claude-3.5-sonnet + T,
  claude-3.5-sonnet DSPy CoT,
  claude-3.5-sonnet DSPy CoT + T,
  claude-3.5-sonnet DSPy ReAct,
  claude-3.5-sonnet DSPy ReAct + T
}
\end{axis}
\end{tikzpicture}
\caption{Comparison of QA performance (Accuracy \%) between the three best performing models, gpt-4o, gpt-4-turbo, and claude-3.5-sonnet, across all languages and methods. {+} T indicates the {+} Translation experiment.}
\end{figure*}

\clearpage

\begin{figure*}[!h]
\label{fig:comprehensive-ner-performance-line}
\captionsetup{width=.8\textwidth}
\centering
\begin{tikzpicture}
\begin{axis}[
    width=15cm,
    height=10cm,
    xlabel={Language},
    ylabel={F1 Score (\%)},
    title={NER Performance: gpt-4o vs gpt-4-turbo vs claude-3.5-sonnet},
    xticklabels={ALS,YO,AZ,TR},
    xtick={1,2,3,4},
    ymin=80,
    ymax=100,
    ymajorgrids=true,
    grid style=dashed,
    legend style={
        at={(0.5,1.25)},
        anchor=south,
        legend columns=3,
        font=\footnotesize
    },
]

\addplot[color=purple,mark=*] coordinates {
    (1,92.01) (2,90.57) (3,89.25) (4,92.23)
};
\addplot[color=purple,mark=square] coordinates {
    (1,91.90) (2,91.04) (3,88.50) (4,90.56)
};
\addplot[color=purple,mark=triangle] coordinates {
    (1,88.32) (2,91.32) (3,90.70) (4,86.88)
};
\addplot[color=purple,mark=diamond] coordinates {
    (1,83.59) (2,85.69) (3,86.05) (4,82.63)
};
\addplot[color=purple,mark=pentagon] coordinates {
    (1,88.32) (2,91.32) (3,90.71) (4,86.80)
};
\addplot[color=purple,mark=star] coordinates {
    (1,83.59) (2,85.69) (3,86.05) (4,82.63)
};

\addplot[color=orange,mark=*] coordinates {
    (1,87.88) (2,85.91) (3,84.62) (4,84.62)
};
\addplot[color=orange,mark=square] coordinates {
    (1,89.78) (2,89.78) (3,87.30) (4,87.96)
};
\addplot[color=orange,mark=triangle] coordinates {
    (1,85.26) (2,90.32) (3,88.10) (4,84.19)
};
\addplot[color=orange,mark=diamond] coordinates {
    (1,85.97) (2,82.83) (3,81.29) (4,84.80)
};
\addplot[color=orange,mark=pentagon] coordinates {
    (1,85.26) (2,90.33) (3,88.11) (4,84.19)
};
\addplot[color=orange,mark=star] coordinates {
    (1,82.83) (2,84.80) (3,85.97) (4,81.29)
};

\addplot[color=teal,mark=*] coordinates {
    (1,93.85) (2,92.61) (3,93.85) (4,89.07)
};
\addplot[color=teal,mark=square] coordinates {
    (1,85.25) (2,87.89) (3,89.28) (4,85.60)
};
\addplot[color=teal,mark=triangle] coordinates {
    (1,92.68) (2,93.21) (3,94.18) (4,88.71)
};
\addplot[color=teal,mark=diamond] coordinates {
    (1,85.04) (2,87.98) (3,86.43) (4,83.38)
};
\addplot[color=teal,mark=pentagon] coordinates {
    (1,87.62) (2,90.70) (3,95.31) (4,88.29)
};
\addplot[color=teal,mark=star] coordinates {
    (1,86.66) (2,89.03) (3,89.11) (4,84.48)
};

\legend{
  gpt-4o,
  gpt-4o + T,
  gpt-4o DSPy CoT,
  gpt-4o DSPy CoT + T,
  gpt-4o DSPy ReAct,
  gpt-4o DSPy ReAct + T,
  gpt-4-turbo,
  gpt-4-turbo + T,
  gpt-4-turbo DSPy CoT,
  gpt-4-turbo DSPy CoT + T,
  gpt-4-turbo DSPy ReAct,
  gpt-4-turbo DSPy ReAct + T,
  claude-3.5-sonnet,
  claude-3.5-sonnet + T,
  claude-3.5-sonnet DSPy CoT,
  claude-3.5-sonnet DSPy CoT + T,
  claude-3.5-sonnet DSPy ReAct,
  claude-3.5-sonnet DSPy ReAct + T
}
\end{axis}
\end{tikzpicture}
\caption{Comparison of NER performance (F1 score) between the three best performing models, gpt-4o, gpt-4-turbo, and claude-3.5-sonnet, across all languages and methods. {+} T indicates the {+} Translation experiment.}
\end{figure*}